\documentclass[10pt, a4paper]{article}
\usepackage[top=3.6cm, bottom=3.2cm, left=3.cm, right=3.cm]{geometry}
\usepackage{mathrsfs}

\usepackage{amssymb}
\usepackage{amsmath}
\usepackage{verbatim}
\usepackage{booktabs}
\usepackage{ctable}
\usepackage{multirow}
\usepackage{multicol}
\usepackage{rotating}
\usepackage{threeparttable}
\usepackage{graphicx}
\usepackage{subfigure}
\usepackage{cite}
\makeatletter

\newcommand{\Rmnum}[1]{\expandafter\@slowromancap\romannumeral #1@}
\makeatother
\usepackage{algorithm}
\usepackage{algpseudocode}

\begin{document}
%
\title{Meaningful Objects Segmentation from SAR Images via A Multi-Scale Non-Local Active Contour Model}

\author{Gui-Song~Xia, Gang Liu, Wen Yang \\
\\
LIESMARS, Wuhan University, Wuhan, China \\
SPL, EIS, Wuhan University, Wuhan, China\\
\{\emph{guisong.xia, gangliu, yangwen}\}@whu.edu.cn
}

\date{}


\maketitle

\begin{abstract}
The segmentation of synthetic aperture radar (SAR) images is a longstanding yet challenging task, not only because of the presence of speckle, but also due to the variations of surface backscattering properties in the images.
Tremendous investigations have been made to eliminate the speckle effects for the segmentation of SAR images, while few work devotes to dealing with the variations of backscattering coefficients in the images.
In order to overcome both the two difficulties, this paper presents a novel SAR image segmentation method by exploiting a multi-scale active contour model based on the non-local processing principle. More precisely, we first formulize the SAR segmentation problem with an active contour model by integrating the non-local interactions between pairs of patches inside and outside the segmented regions.
Secondly, a multi-scale strategy is proposed to speed up the non-local active contour segmentation procedure and to avoid falling into local minimum for achieving more accurate segmentation results.
Experimental results on simulated and real SAR images demonstrate the efficiency and feasibility of the proposed method: it can not only achieve precise segmentations for images with heavy speckles and non-local intensity variations, but also can be used for SAR images from different types of sensors.


\end{abstract}



%

\section{Introduction}

Synthetic aperture radar (SAR) systems have been widely used in remote-sensing applications for many years because of their advantages of long-distance performance, strong penetrability and all-weather acquisition capability. However, the understanding of SAR data is a long-standing and challenging task, due to the presence and the strong multiplicative nature of speckle noises~\cite{maitre2010processing,Soumekh}. This article addresses the segmentation of SAR images, which is an active topic and one of the fundamental problems for the interpretation of SAR images~\cite{maitre2010processing,Deng2005,Marques,Xia,Shuai,Li2010,feng2013multiphase}.

\subsection{Motivation and objective}
Image segmentation is regarded as a process of partitioning an image into homogeneous regions with respect to certain computable visual characteristics. By grouping image pixels into segments, the goal of image segmentation is usually simplifying an image and establishing a compact and meaningful image representation for the subsequent analysis. One should notice that, in a general setup, image segmentation is a goal-dependent, subjective, and hence ill-posed problem, which is extremely intractable to find a generic solution~\cite{Sumengen2006}. Deep discussions on image segmentation could be found in~\cite{Morel1995,Shapiro}.

In the context of SAR image analysis, roughly speaking image segmentation is usually used for two main tasks: for achieving an over segmentation of a SAR image which can be taken as a mid-level representation in subsequent processes (e.g. object-oriented classification), and extracting/segmenting out meaningful objects (e.g. oil spill regions) from a given image.
\begin{figure}[ht!]
\centering
  \subfigure[]{\includegraphics[height= 0.32\linewidth]{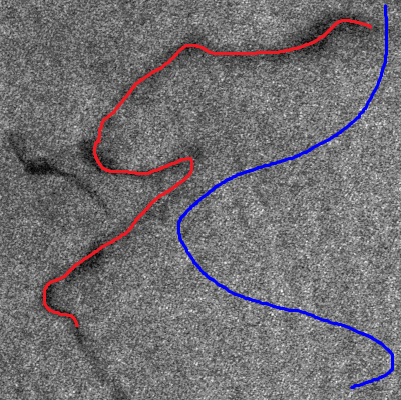}}
  \subfigure[]{\includegraphics[height= 0.32\linewidth]{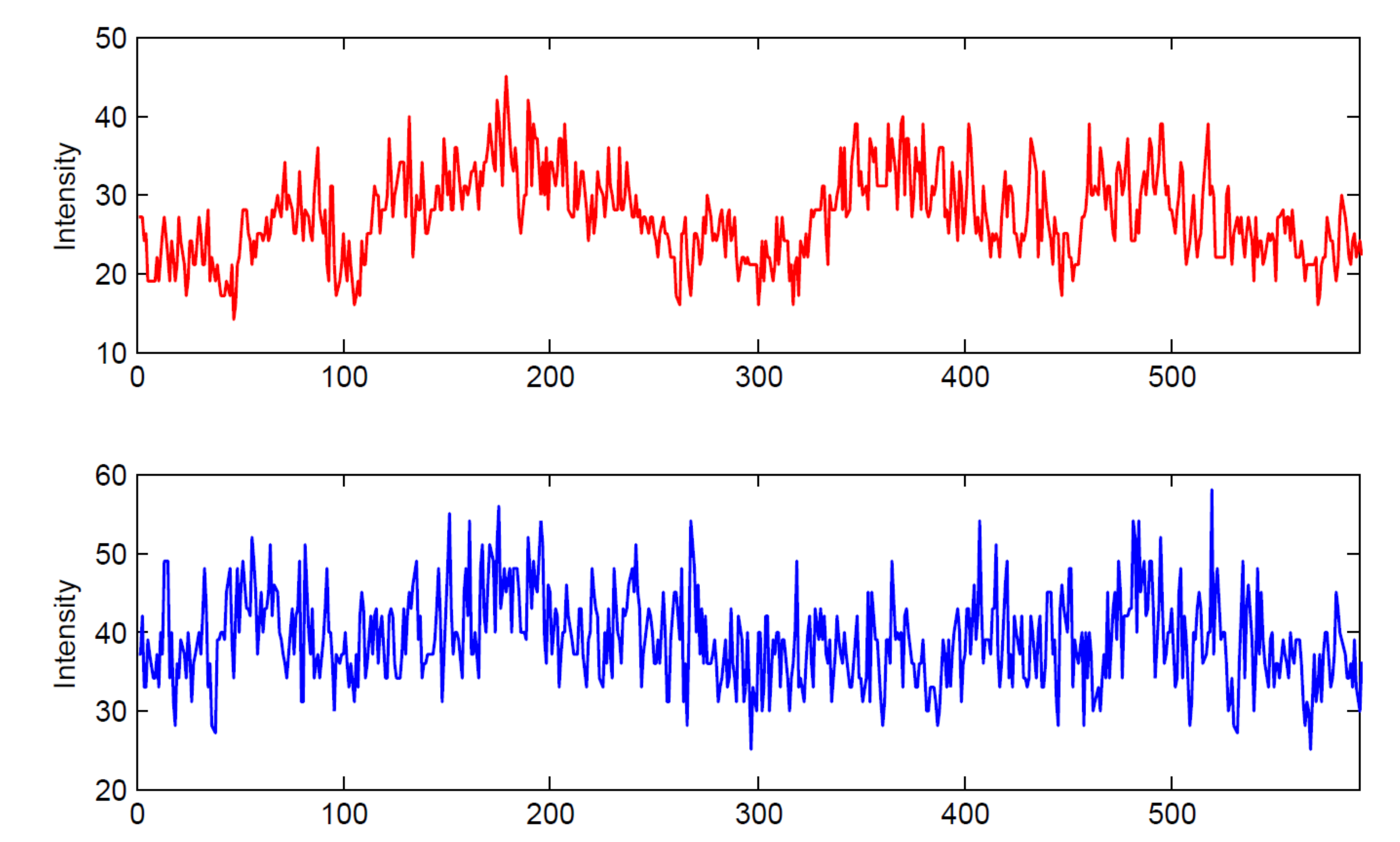}}
  \caption{The difficulties in the SAR image segmentation task. For the sake of illustration, (a) shows two profiles, annotated by curves in different color overlayed on the original image, of a C-band ENVISAT ASAR image. The red curve is drawn on the object, while the blue one is located on the background. (b) displays the two one-dimensional profiles as 1-D signals, correspondingly. One can note the multiplicative nature of the speckle noises and the non-local variations of backscattering properties in the image, which affect the segmentation results a lot, as we shall see in experiments. Refer to the texts for more explanations. }
  \label{fig:SAR_profiles}
\end{figure}

This paper particularly focuses on the problem of meaningful objects segmentation from SAR images, which is difficult mainly due to following two facts:
\begin{itemize}
\item[-] \emph{The speckle effects in SAR images}. As illustrated in Figure~\ref{fig:SAR_profiles}, the multiplicative nature of the speckle noise in SAR images is undesired for image segmentation, which often leads to small isolated regions in the segmentation results. In order to eliminate these effects, efficient stochastic models of speckles are usually employed~\cite{gao2010statistical} and regularization constraint is also imposed to the segmentation boundaries~\cite{Deng2005,Xia,Shuai,feng2013multiphase}.
\item[-] \emph{The non-local variations of surface backscattering properties in the backgrounds or objects}. In a single polarization radar  image, often displayed as a grey scale image, the intensity of each pixel represents the proportion of microwave backscattered from that area on the ground which depends on a variety of factors\footnote{The pixel intensity values of a SAR image are often converted to a physical quantity called the backscattering coefficients, which usually depends on: types, sizes, shapes and orientations of the scatterers in the target area; moisture content of the target area; frequency and polarisation of the radar pulses; as well as the incident angles of the radar beam. }. In the context of extracting objects from SAR images, the backscattering coefficients of background and objects are actually supposed to be homogeneous, respectively. However, in the case of low- or middle-resolution SAR images, where each pixel covers a large area on the surface, there is no guarantee that the interested regions are homogeneous. For instance, as shown in Figure~\ref{fig:SAR_profiles}, non-local changes of image pixel values appear both in the background and objects. As we shall see, these variations might be slight, but they affect the segmentation results a lot and it is intricate to solve this problem by imposing more regularization to the segmentation.
\end{itemize}

In the literature, tremendous studies have been devoted to solving the first problem in SAR image segmentation, e.g.~\cite{Fjortoft1999,Deng2005,Li2010,Zribi,Zhang,Xia,Bombrun2011,Tello2011,Shuai,feng2013multiphase,Marques}, among which active contour models show  many advantages and promising results at finding accurate object boundaries in the presence of speckles. While, to our knowledge, few method can deal with the non-local variations of pixel values (i.e. backscattering coefficients) in SAR images. In the past few years, it has been reported that non-local principles~\cite{Del,Deledalle2015} can achieve much better de-speckling results than that of using local stochastic models of speckles. Meanwhile, Jung {\em et al.}~\cite{Jung} has currently presented that an active contour model with the non-local principle showed superiority on segmenting natural images with texture variations. It is thus of great interest to investigate the use of non-local principle for SAR image segmentation, combining with active contour models.


\subsection{Related Work}
\subsubsection{Active contour methods for SAR image segmentation}
Active contour model is a framework for delineating the outlines of object from images, by minimizing an energy function associated to a contour as a sum of two terms: the \emph{internal energy} that reflects the properties of the contour itself and the \emph{external energy} that related to the objects~\cite{Kass}. Since the model shows good performance on object segmentation from noisy images, it has been widely studied in SAR image segmentation.
Early works using active contour models on the problem of SAR image segmentation include~\cite{horritt1999statistical,Germain}, where investigations were mainly devoted to designing the two-term energy function and to evolve the contours in SAR images.
For instance, Horritt~\cite{horritt1999statistical} proposed to measure local profiles, e.g. local tone and texture, along the contour and pursue a final object boundary under a curvature constrain. While, in~\cite{Germain}, Germain {\em et al.} used a likelihood-ratio edge detector to localize edges in images and employed the ``snake model"~\cite{Kass} to refine edges subsequently to achieve a final segmentation. One difficulty of these works, however, lies in the fact that the convergence of discrete ``snake'' curves is usually with high computational complexity.

Among all the methods that attempts to speedup the evolution of contours, level-set based implementation~\cite{Malladi1995,Osher} is very efficient and has been widely used for SAR image segmentation, see~\cite{ayed2005multiregion,Shuai,Hu,feng2013multiphase}.
Specifically, Martin {\em et al.}~\cite{martin2004influence} studied the influence of noise models to the level-set active contour for image segmentation and demonstrated that this kind of method is robust to noises which can be described by a model in exponential family.
Ayed \emph{et al.} proposed an algorithm to evolve simple closed planar curves with an explicit correspondence between the interiors of curves and segmented regions~\cite{ayed2005multiregion}.
Using Gamma distributions for homogeneous regions in SAR images, \cite{Shuai} proposed a level-set energy functional to get a stationary global minimum for segmentation.
While, Hu\cite{Hu} proposed to use Kullback-Leibler distance of Edgeworth to accurately segment SAR images.
Feng {\em at al.}~\cite{feng2013multiphase} presented a variational multi-phase method for SAR image segmentation.

Observe that most of these methods attempted to minimize the energy inside and outside the segmented regions, based on the hypothesis that pixels inside and outside the segmented regions are consistent and follow the same distribution respectively~\cite{Ayed2006,Mejail2008,lu2009active}, which is however not true for many real SAR images.
In contrast, Marques~{\em et. al.}\cite{Marques} explored statistical properties of SAR data to characterize image regions and derive an energy functional to perform region mapping, which splits SAR images into homogeneous, heterogeneous and extreme heterogeneous regions. This method reported state-of-the-art segmentation results for most SAR images, such as SAR images with less textures and speckle effects.
However, it is still not clear how to segment SAR images with non-local radiometric variations in the images.

\subsubsection{Non-local active contours for image segmentation}
The idea of using non-local active contour model (NLAC) for natural image segmentation is first proposed by Jung {\em et al.}~\cite{Jung}, which, in contrast to other active contour methods, only requires the local homogeneity between patches.
Particularly, a non-local energy is defined by using a pairwise interaction of patch features inside and outside the segmented regions. The pairwise interaction of patch features are calculated by measuring the patch similarity in a non-local way~\cite{Buades}. The NLAC method has shown promising results on natural image segmentation, however the convergence of the method is slow, meaning that it will be time consuming to achieve a satisfied segmentation results. Moreover, how the method can be adapted to SAR images still remains to be a problem.

\subsection{Contributions of this work}


Inspired by the work of Jung {\em et al.}~\cite{Jung} for natural images, this paper therefore presents a novel SAR image segmentation method by relying on a multi-scale active contour model with non-local processing principle. More precisely, this method calculates the patch similarity in the SAR image in a non-local manner, and formalizes the segmentation problem with a multi-scale active contour model by integrating the interactions between pairs of patches inside and outside the segmented region. Particularly, the SAR image is decomposed into several scales. At each scale, the segmented result of previous scale is used to initialize the current input. In our multi-scale model, the contour propagates more accurately and quickly.

Our work is distinguished by three major contributions:
\begin{itemize}
\item[-] We introduce the non-local active contour model for SAR image segmentation, by investigating different pair-wise patch similarities for computing the non-local energy. To our knowledge, it is the first time to adapt the non-local active contour model for the problem of SAR image segmentation.
\item[-] We propose a multi-scale strategy, the multi-scale NLAC method, to speed up the convergence of the NLAC model. With this strategy, the segmentation result of previous scale is used to initialize the NLAC model at current scale. Thus, at each scale, what the algorithm does is to refine the segmentation result at coarse scale to obtain a finer and more accurate result with the NLAC algorithm. This strategy can help the NLAC model to avoid trapping into local minimum but quickly converge to a better solution. As a general setting, this multi-scale NLAC method can actually be appropriate for segmenting other type of images as well.
\item[-] Our method can be well applied to segmenting SAR images from various types of sensors (meaning different speckle models), as the region homogeneity is measured by means of integrating the interactions between pairs of patches, \emph{i.e.} patch similarity, rather than using the probabilistic models of SAR images directly. We also evaluate the patch similarity with different SAR image speckle models and metrics.
\end{itemize}

The remainder of the paper is organized as follows: In Section~\Rmnum{2}, we review the background on active contour model for SAR image segmentation. In Section~\Rmnum{3}, we present the basic algorithm of non-local active contour model for SAR image segmentation. In Section~\Rmnum{4}, we explain the proposed multi-scale strategy for non-local active contour model for SAR image segmentation. We test our algorithm in Section~\Rmnum{5} on both synthetic and real SAR images. Finally, we conclude our work in Section~\Rmnum{6}.

\section{Background on active contour model for SAR image segmentation}
Let $f:\Omega \rightarrow R$ be a SAR image with $\Omega$ as the image grid. The goal of SAR image segmentation is to partition the image domain $\Omega$ into $N$ pairwise disjoint regions $\{\mathcal{R}_i\}_{i=1}^N$ with respect to certain statistical characteristics of homogeneity, where the regions $\{\mathcal{R}_i\}_{i=1}^N$ satisfy the constraints $\mathcal{R}_i \bigcap \mathcal{R}_j|_{\forall i\neq j}=\emptyset$ and $\bigcup_{i=1}^N \mathcal{R}_i=\Omega$ for all $i,j \in \{1,\ldots, N]\}$.

With the assumption that SAR image data $\{z | z = f(s), s \in \Omega \}$ can be described by a probabilistic distribution (\emph{e.g.} Gamma distribution, $G_A^0$ distribution) $P_\mathbf{\theta}(z)$, with $\mathbf{\theta} = \{\theta_i\}_{i=1}^N$ as the parameter vector of the distribution, the segmentation $\{\mathcal{R}_i\}_{i=1}^N$ of the SAR image $f$ can be achieved by maximizing the log-likelihood
\begin{align}
\hat{\mathcal{R}} = \arg\max_{\mathcal{R}} \sum_{i=1}^N\sum_{s\in \mathcal{R}_i} \log \big(P_{\mathbf{\theta_i}} ( f(s) ) \big).
\label{eq:MLE}
\end{align}

Because of the disturbance of the speckle in SAR image, this method obtains a segmentation of SAR image with many isolated small regions, which, however, can be avoided by adding a regularizing term $E_R$ to the log-likelihood energy $E_\Theta$. The regularizing term is often defined as the length of the contours~\cite{zhu1996region,feng2013multiphase}. Subsequently, a new segmentation can be achieved by minimizing following energy,
\begin{align}
E = - \underbrace{\sum_{i=1}^N \int_{\mathcal{R}_i} \log(P_{\mathbf{\theta_i}}(f(s)))ds}_{E_\Theta} + \lambda \, \underbrace{\sum_{i=1}^{N}\int_{\partial \mathcal{R}_i} g(s) \,ds }_{E_R},
\label{eq:e0}
\end{align}
where $g(s)=\frac{1}{1+|G_\sigma\ast\nabla f|^2}$ denotes the boundary indicator function that includes the information of edges with $G_\sigma$ denoting a Gaussian kernel with scale $\sigma$ and $\ast$ denoting the convolution symbol.
$\partial\mathcal{R}_i, i\in \{ 1,2,\ldots,N\}$ denotes the contour of region $\mathcal{R}_n$ and $\lambda$ is a weighted parameter used to balance these two terms.

It is worth noticing that this conventional active contour model is mainly based on the hypothesis that the pixels in each segmented region of SAR image are piecewise consistent and follow the same probabilistic model. It performs well for many SAR image partitioning and investigations have been focused on pursuing good probabilistic distributions and global minimization of the energy~\cite{Shuai}. This type of methods, however, can not well sperate the background and objects which can not be modeled by using only one probabilistic distribution, as illustrated in Figure~\ref{fig:SAR_profiles}.

\section{Non-local active contour model for SAR image segmentation}

Non-local image processing refers to the general methodology of designing energies using nonlocal comparisons of patches extracted in the image~\cite{Buades2005}, which have proven to be efficient for many imaging problems~\cite{buades2008nonlocal,arias2011variational,deledalle2011nl}.
Recently, Jung {\em et al.}~\cite{Jung} have used a non-local energy that enforces the nonlocal similarity of patches inside each region to be segmented for image segmentation. This patch comparison principle drives the active contour to optimize the homogeneity of each region.

In what follows, we first briefly recall the non-local active contour model and then adapt it to SAR image segmentation.

\subsection{Non-local active contour model}
The main idea of non-local active contour (NLAC) is to minimize the non-local energy which is defined by integrating the interactions between pairs of patches inside and outside the segmented region with a level set method.

For a given image $f$, let $p_s(k)=f(s+k), k \in \{-\tau, \tau\}^2$ be a patch around the location $s \in \Omega$ with size of $(2\tau + 1)^2$. The NLAC model calculates the segmentation of the image $f$ by minimizing the energy,
\begin{align}
\label{energy}
E(\mathcal{R}) = E_D(\mathcal{R}) + \lambda \, E_R(\mathcal{R})
\end{align}
where $E_D(\mathcal{R})$ is an energy term associated with the image data, measuring the dissimilarity inside and outside the region $\mathcal{R}$. $E_R(\mathcal{R})$ is a regularization term imposed to the region $\mathcal{R}$ and $\lambda>0$ is a weight to balance the two terms.

\begin{itemize}
\item[-] {\textbf{Data term}} $E_D(\mathcal{R})$ is defined as
\begin{align}
\label{eq:E1}
E_D(\mathcal{R}) = \bar{E}_D(\mathcal{R}) + \bar{E}_D(\mathcal{R}^c),
\end{align}
where $\mathcal{R}^c = \Omega / \mathcal{R}$ is the complementary of the region $\mathcal{R}$ and
$$
\bar{E}_D(\mathcal{R}) = \int_{\mathcal{R} \times \mathcal{R}} G_{\sigma}(s, t) d(p_s, p_t)\, ds\, dt,
$$
with $G_{\sigma}(s, t)$ as a Gaussian kernel of scale $\sigma$. The term $d(p_s, p_t)$ measures the dissimilarity or distance between the two patches $p_s$ and $p_t$.
Here a specific distribution is used to model the patches and metrics between the distribution of patches are calculated to obtain the distance $d(p_s, p_t)$, which will be elaborated later.
The first term $\bar{E}_D(\mathcal{R})$ measures the internal energy while the second term $\bar{E}_D(\mathcal{R}^c)$ measures the external energy.

It has been proved that the region $\mathcal{R}$ capturing the objects of interest can be evolved by the means of level-set function, which is defined as
\begin{equation}
\begin{cases}
\varphi(s)>0,\quad \text{if} \; s\in \mathcal{R}\\
\varphi(s)=0,\quad \text{if} \; s \in\partial \mathcal{R}\\
\varphi(s)<0,\quad \text{if} \; s \not\in \mathcal{R}.
\end{cases}
\end{equation}
Thus, we can rewrite Eq.~\eqref{eq:E1} as
\begin{align}
&E_D(\mathcal{R}) = \varepsilon_D(\varphi) \nonumber \\
&= \int_{\Omega \times \Omega} \rho \Big( H\big(\varphi(s)\big), H\big(\varphi(t)\big) \Big)
          G_{\sigma}(s, t) d(p_s, p_t) \,ds \,dt
\label{eq:E2}
\end{align}
where $H(u)= \frac{1}{2}+\frac{1}{\pi}\arctan(\frac{u}{\epsilon})$ is the Heaviside function
with the parameter $\epsilon$ small enough for achieving a sharp boundary at the point $u=0$.
It is obvious that $H(u)=1$ when $u\leq 0$ and $H(u)=0$ when $u>0$.
$\rho(u,v)=1-|u-v|$ is an indicator function implying that only pairs of pixels for which $\varphi$ has the same sign are considered. This function calculates the difference of energy inside and outside the segmented region.

\item[-] \textbf{Regularization term} For the purpose of regularizing the extracted region, the segmented region is penalized by the length of the contour $C$ as
\begin{align}
\label{Lfunction}
{E}_R(\mathcal{R}) &= \varepsilon_R(\varphi) = \int\|\nabla H(\varphi(s))\| \,ds \nonumber \\
                       &= \int H'(\varphi(s))\Vert\nabla\varphi(s)\Vert \, ds,
\end{align}
where $\nabla H(\varphi(s))$ indicates the gradient at the point $s$ of the function $H(\varphi)$.
\end{itemize}

So far, we have the non-local active contour energy, and one can use an iterative scheme proposed in~\cite{Jung} to obtain the segmentation by relying on the gradient descent algorithm with an artificial time $t\geq 0$, which leads to the evolution equation for $\varphi$,
\begin{equation}
\label{eq:varphi_t}
\frac{\partial\varphi}{\partial t} = -(\nabla \varepsilon(\varphi)) = -\left(\nabla \varepsilon_D(\varphi) + \lambda \nabla \varepsilon_R(\varphi) \right),
\end{equation}
where the two terms are calculated as
\begin{align}
\label{eq:e}
\nabla \varepsilon_D(\varphi)(s) &=\int (\partial \rho)\left(H(\varphi(s)), H(\varphi(t))\right) \nonumber \\
                     &\cdot G_\sigma(s-t)\cdot d(p_s, p_t) \,dt\cdot H'(\varphi(s)),
\end{align}
and
\begin{align}
\label{eq:l}
\nabla \varepsilon_R(\varphi)(s)= -div\left(\frac{\nabla\varphi(s)}{\Vert \nabla\varphi(s)\Vert}\right)H'(\varphi(s)).
\end{align}
We discretize the iterative scheme and obtain
\begin{align}
\label{eq:varphi_i}
\varphi^{(i+1)}=\varphi^{(i)}-\xi\cdot(\nabla \varepsilon_D(\varphi) + \lambda \nabla \varepsilon_R(\varphi) ),
\end{align}
where $\xi$ is the time step size. The iteration of this method stops if $\vert \varepsilon^{(i)}- \varepsilon^{(i+1)}\vert<\ \omega$, where $\omega$ is a pre-defined threshold.

\begin{table*}[htb!]
\centering
\caption{Different probabilistic distributions $\mathbb{P}(z)$ for characterizing local SAR images, where $E[x]$ and $\textrm{Var}[x]$ are denoted as the expectation and variance of $z$ respectively.}
\label{tab:distribution}
\begin{tabular}{c|l|l}
\hline
 Distribution & Mathematical representation & Estimation of parameters \\  \hline
 Log-normal   & $\mathbb{P}(z; \mu, \sigma)=\frac{1}{\sqrt{2\pi}\sigma z} e^{- (\ln z - \mu)^2/{2\sigma^2}}$ &
             $ \mu = \ln \frac{E^2[z]}{\sqrt{\textrm{Var}[z]+E^2[z]}} $, \,
             $\sigma^2 = \ln \big( \frac{\textrm{Var}[z]}{E^2[z]}+1\big)$.
             \\ \hline
 Rayleigh &  $\mathbb{P}(z; \sigma)=\frac{z}{\sigma^2} e^{- z^2/{2\sigma^2}}$ & $\sigma^2=\frac{2}{4-\pi} \textrm{Var}[z]$.
                          \\ \hline
 Gamma & $\mathbb{P}(z; \alpha, \beta)=\frac{\beta^\alpha}{\Gamma(\alpha)} z^{\alpha-1} e^{-\beta z}$ &
         $\alpha = {E^2[z]}/{\textrm{Var}[z]}$, \, $\beta= {E[z]}/{\textrm{Var}[z]}$.
          \\ \hline
 Weibull & $\mathbb{P}(z; \beta, \eta) = \frac{\beta z^{\beta-1}}{\eta^{\beta}} e^{ -z^{\beta}/{\eta^{\beta}} }$ &
         $\begin{cases}
        \textrm{Var}[z]/{E^2[z]} = \Gamma(1 + \frac{2}{\beta})/\Gamma^2(1 + \frac{1}{\beta}) \\
        \eta = E^{\beta}[z]/{\Gamma^{\beta}(1+ \frac{1}{\beta})}
        \end{cases}$
        \\ \hline
 $G_A^0$ & $\mathbb{P}(z; \alpha, \gamma)=\frac{2n^n\Gamma(n-\alpha)}{\gamma^\alpha\Gamma(-\alpha)\Gamma(n)}\frac{z^{2n-1}}{(\gamma+z^2n)^{n-\alpha}}$ &
    $
    \begin{cases}
    \frac{\Gamma^2(-\alpha-\frac{1}{4})}{\Gamma(-\alpha)\Gamma(-\alpha-\frac{1}{2})}-\frac{m^2_{{1}/{2}}}{m_1}\frac{\Gamma(n)\Gamma(n+\frac{1}{2})}{\Gamma^2(n+\frac{1}{4})}=0 \\
    \gamma = E^2[z]n \cdot (\frac{\Gamma(-\alpha)\Gamma(n)}{\Gamma(-\alpha- \frac{1}{2})\Gamma(n+ \frac{1}{2})})^2
    \end{cases}
    $
          \\ \hline
\end{tabular}
\end{table*}

\begin{table*}[htb!]
\centering
\caption{Different dissimilarity measures used for comparing the pairwise local patches of SAR images. Let $P = \{P_j \}_{j=1}^N$ and $Q= \{Q_j\}_{j=1}^N$ be two probabilistic mass distributions (histograms in experiments), $D(P \| Q)$ gives the dissimilarity measures between them.}
\label{tab:distance}
\begin{tabular}{c|l}
\hline
 Dissimilarity measures & numerical computation of $D(P \| Q)$ \\  \hline
 Kullback-Leibler (KL)   & $D_{KL}(P \| Q) = \sum_{j=1}^N P_j\cdot \ln \frac{P_j}{Q_j} + Q_j\cdot \ln \frac{Q_j}{P_j}$.  \\ \hline
 Hellinger   & $D_H (P \| Q) = \frac{1}{\sqrt{2}} \big(\sum_{j=1}^N \big(\sqrt{P_j} - \sqrt{Q_j}\big)^2 \big)^{\frac{1}{2}}$. \\ \hline
 Total Variation & $D_{TV} (P \| Q) = \frac{1}{2} \sum_{j=1}^N \big| P_j - Q_j \big|$.  \\ \hline
 Jensen-Shannon (JS) & $D_{JS} (P \| Q) = \sum_{j=1}^N P_j\cdot \ln \frac{P_j}{P_j + Q_j} +Q_j\cdot \ln \frac{Q_j}{P_j+Q_j}$. \\ \hline
 Earth Mover (EM) distance & $D_{EM} (P \| Q)=\sum_{j=1}^N \big( \sum_{i=1}^j P_j-\sum_{i=1}^j Q_j \big)$. \\ \hline
\end{tabular}
\end{table*}

\subsection{SAR image segmentation by comparing local patches}
As mentioned above, the basic hypothesis of SAR image segmentation with non-local active contour models is that SAR images are piecewise homogeneous in the object and background regions respectively, thus the segmentation can be implemented by measuring the similarities between pairwise local patches.
In order to calculate the patch similarities, we use the statistical distributions for describing patches and compute the distance between such distributions.

In the proposed segmentation algorithms, local patch similarities, $d(p_s, p_t)$ defined above, play an important role in the computation of non-local energy.
The similarities of pairwise patch have been recently investigated by Deledalle~\emph{et al.}~\cite{Del,Deledalle2015}, showing that the probabilistic characteristics~\cite{Soumekh} perform the best among others on SAR images.


More precisely, let $z$ be the intensity of the patch $p_s$ in a SAR image, we denote $\mathbb{P}(z)$ as the probabilistic mass function of $z$. In this work, we investigate different probabilistic distributions, \emph{e.g.} log-normal distribution, Rayleigh distribution, gamma distribution, Weibull distribution and $G_A^0$-distribution, as shown in Table~\ref{tab:distribution} for evaluating NLAC-based SAR image segmentation.
Once we have a histogram-like feature for each patch, we also study a small set of distances between PMFs for measuring dissimilarity, as shown in Table~\ref{tab:distance}.

The detail of the NLAC algorithm is given by Algorithm~\ref{alg:NLAC}.

\begin{algorithm}[htb!]
 \caption{NLAC algorithm for SAR image segmentation}
 \label{alg:NLAC}
 \begin{algorithmic}[1]
 \Require  {a SAR image $f$, initialization $\varphi_0$, weight $\lambda$, iteration-stop thresholds $\omega$;
 }
 \Ensure  {a segmentation $S$}
  \State Initialization: $i \leftarrow 0$; $\varphi^{(0)} \leftarrow \varphi_0$;
  \While{$\vert \varepsilon^{(i+1)}-\varepsilon^{(i)}\vert < \omega$}
    \State Computing local patch dissimilarities by choosing a feature in Table~\ref{tab:distribution} and a similarity measure in Table~\ref{tab:distance};
    \State Computing the energy gradient with Eq.~\eqref{eq:varphi_t},\eqref{eq:e},\eqref{eq:l};
    \State Updating $\varphi^{(i+1)}$ with $\varphi^{(i)}$ by Eq.~\eqref{eq:varphi_i};
    \State $i \leftarrow (i+1)$;
  \EndWhile
  \State $S\leftarrow H(\varphi)$.
 \end{algorithmic}
\end{algorithm}

\section{Multiscale Non-local active contour model for SAR image segmentation}

One disadvantage of the NLAC algorithm is its low speed of convergence. Given SAR image with size of $ M\times N$, in each interaction, one needs to calculate the non-local energy of every pixel with a complexity of \textit{$M^2N^2$}, which makes the algorithm slow.
Another problem of the NLAC algorithm lies in the size selection of local patches: big patches provide more robust and better estimation of the probabilistic model, while result in segmentations with lower precision as they often lead to blurred segmentation boundaries if the patches across different regions.

\begin{figure*}[htb!]
\centering
  \includegraphics[width= 0.85 \linewidth]{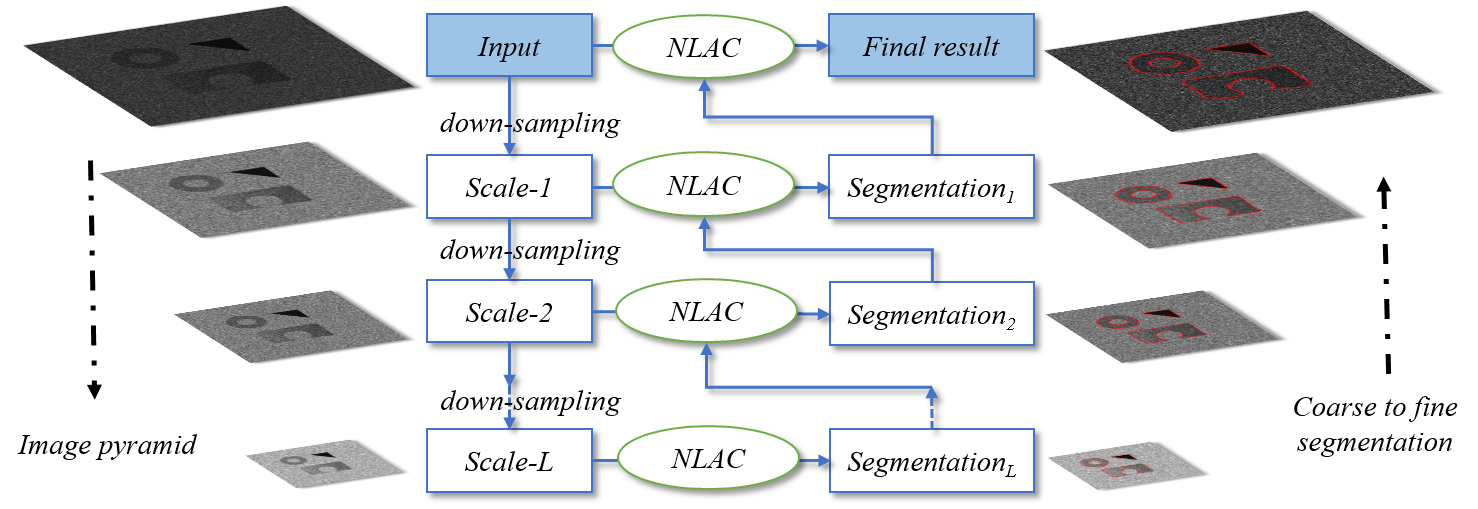}
  \caption{The flowchart of the proposed method. The input SAR image is down-sampled successively to form an image pyramid with $L$ scales, with which we partition the SAR image from coarse to fine with the NLAC method. For an image on the pyramid at scale-$\ell$, on one side, it is segmented by the NLAC algorithm with an initialization from the segmentation of the image at the coarser scale-$(\ell+1)$; on the other side, its result is then up-sampled and fed to the partitioning process of the image at the finer scale-$(\ell-1)$ as an initialization. Note that, for the segmentation of the image at the coarsest scale-$L$, the NLAC algorithm is initialized randomly.}
  \label{fig:flowchart}
\end{figure*}

In order to overcome these drawbacks, we derive the non-local active contour (NLAC) model under a multi-scale strategy. The flowchart of our method is illustrated in Fig.~\ref{fig:flowchart} :

\begin{itemize}
\item[-] \emph{Building the image pyramid} : The input SAR image $f$ with size of $M \times N$ is down-sampled successively to form an image pyramid of $L$ scales,
    $$
    \mathcal{P}(f) = \Big\{ f^{[0]}, f^{[1]}, \ldots, f^{[L-1]} \Big\},
    $$
    where $L$ is an integer between $1$ and $\log_2 \min(M, N)$ and $f^{[0]} \doteq f$ is the original image. $f^{[\ell]}$ is the image at scale-$\ell$ on the pyramid $\mathcal{P}(f)$ computed as
    $$
    f^{[\ell]} = D_{\downarrow \times 2}(G_{\sigma_0} \star f^{[\ell-1]}), \, \forall \ell \in \{1, \ldots, L-1\},
    $$
    where $G_{\sigma}$ is a spatial Gaussian kernel with scale $\sigma_0$, and $D_{\downarrow \times 2}(\cdot)$ is an image down-sampling operator which draws samples in the image domain with spatial step size as $2$, \emph{i.e.}
    $$
    D_{\downarrow \times 2}(f)(x, y) = f(2x, 2y).
    $$

\item[-] \emph{NLAC segmentation on the pyramid} : Based on the image pyramid $\mathcal{P}(f)$, we develop a coarse-to-fine segmentation algorithm of the SAR image $f$. Specifically, for the image $f^{[L-1]}$ at the coarsest scale on the pyramid $\mathcal{P}(f)$, we apply the NLAC algorithm to get a segmentation with random initialization; for the image $f^{[\ell]}$ with $\ell < L-1$, on one side, it is segmented by the NLAC algorithm with an initialization from the segmentation of $f^{[\ell+1]}$ the image at the coarser scale-$(\ell+1)$; on the other side, its result is then up-sampled and fed to the partitioning process of the image at the finer scale-$(\ell-1)$ as an initialization. Finally, The final segmentation of image $f$ is the result achieved on the image at the finest scale.
\end{itemize}


Thus, the principal steps of the proposed MS-NLAC segmentation method are summarized in Algorithm~\ref{alg:MS-NLAC}.

\begin{algorithm}[htb!]
 \caption{MS-NLAC for SAR image segmentation}
 \label{alg:MSNLAC}
 \begin{algorithmic}[1]
 \Require  {a SAR image $f$, weight $\lambda$, thresholds $\omega$, number of scales $L$; }
 \Ensure  {a segmentation $S$}
  \State Initialization: $\varphi^{(L)} \leftarrow \textrm{random initialization}$, $\ell \leftarrow L$;
  \State Build the image pyramid $\mathcal{P} = \{f^{[1]}, f^{[2]}, \ldots, f^{[L]}\}$;
  \While{$\ell \geq 0$}
    \State Updating $\varphi^{(\ell - 1)} \leftarrow NLAC(f^{[\ell]}, \varphi^{(\ell)}, \lambda, \omega)$;
    \State $\ell \leftarrow (\ell-1)$.
  \EndWhile
  \State $S\leftarrow H(\varphi^{(0)})$
 \end{algorithmic}
\end{algorithm}


\section{Experimental results and Analysis}
This section evaluates the proposed method on both simulated and real SAR images. We first describe our experiment setups: the data, the setting of parameters and the quantitative evaluations. Then we analyze the necessity of non-local active contour model by providing an example in which conventional  active contour model can not solve. Next, we demonstrate in detail how the multi-scale strategy can help the NLAC method with a simulated SAR image. Finally, we test our method on different SAR images. The proposed method is also compared with those in~\cite{Shuai} and~\cite{Marques}, which are supposed to be the state-of-the-art.

\subsection{Experiment setups}


\subsubsection{Setting of parameters on the pyramid}
In our method, the patch size $\tau$, the non-local size $\sigma$, and the weight $\lambda$ that balances the fidelity term and the penalty term are parameters needed to be set by users.
In all our experiments, these parameters at different scales on the multi-scale pyramid remains the same. It makes sense that we can look into a larger scope of statistics with the same non-local size parameter, as the scale increase.
We provide the setting of all these parameters for each experiment in what follows.

\subsubsection{Quantitative evaluations} In order to evaluate the segmentation performance quantitatively, we take the same measurement as that used in~\cite{Marques}, where the accuracy of a segmentation $\mathcal{R}$ is measured by the \emph{region fitting error} (RFE), according to a referred segmentation groundtruth $\mathcal{R}_g$, defined as
\begin{align}
 \textrm{RFE}(\mathcal{R}) = \frac{ | \mathcal{R} \bigcup \mathcal{R}_g |_{\textrm{card}} - | \mathcal{R} \bigcap \mathcal{R}_g |_{\textrm{card}}}{ | \mathcal{R}_g |_{\textrm{card}}},
 \label{eq:accurate}
\end{align}
where $|\cdot|_{\textrm{card}}$ is an operator computing the cardinal number of a set. Obviously, the smaller the RFE value is, the better the results are. Especially, RFE equals zero indicating that the segmentation results match the reference perfectly.

\subsection{Are non-local active contour models necessary?}
\begin{figure*}[htb!]
\centering
  \includegraphics[width=0.24\textwidth]{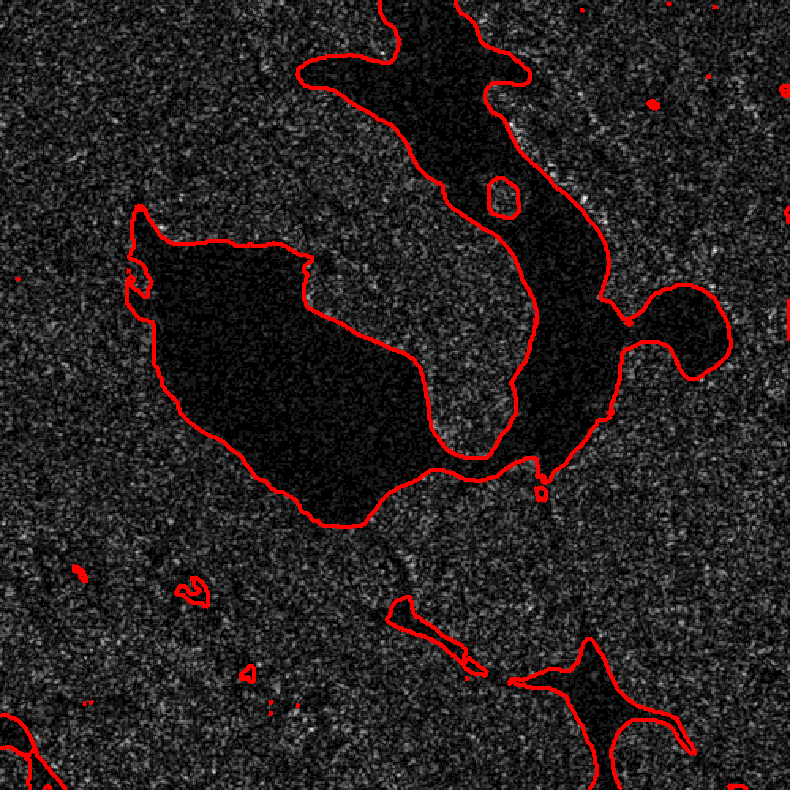}
  \includegraphics[width=0.24\textwidth]{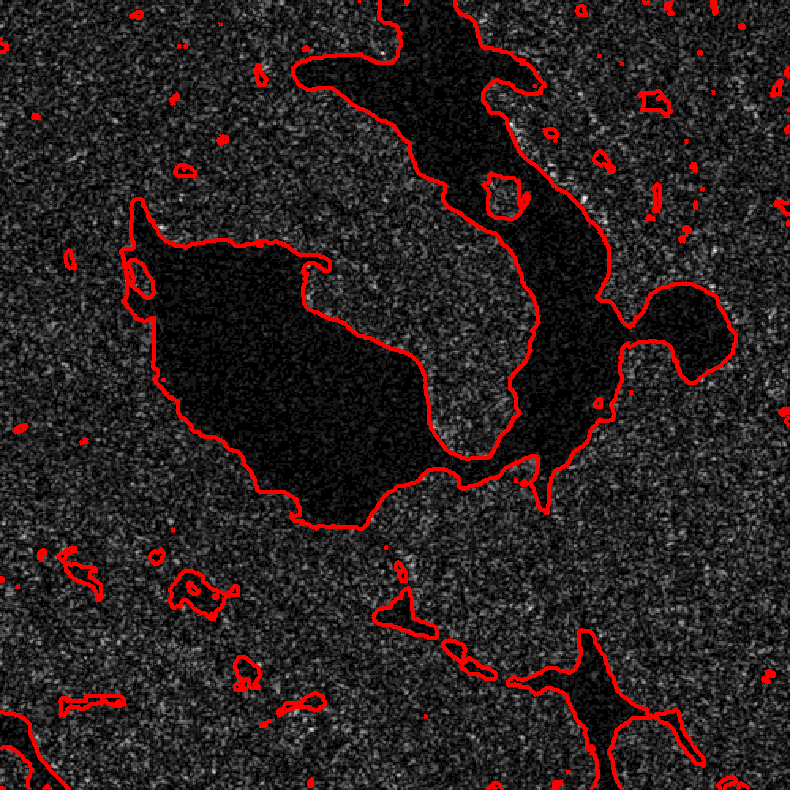}
  \includegraphics[width=0.24\textwidth]{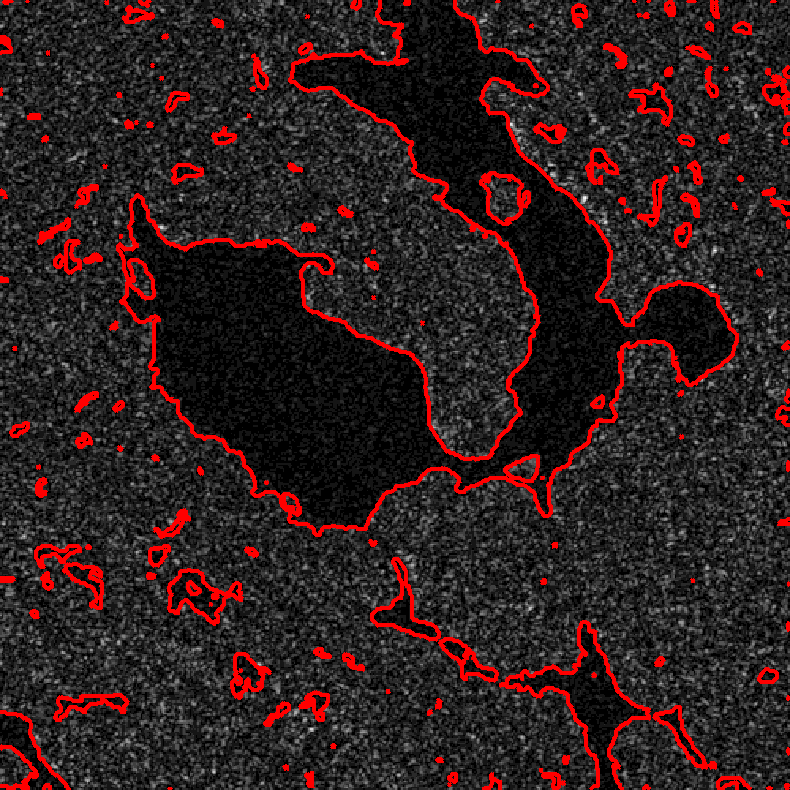}
  \includegraphics[width=0.24\textwidth]{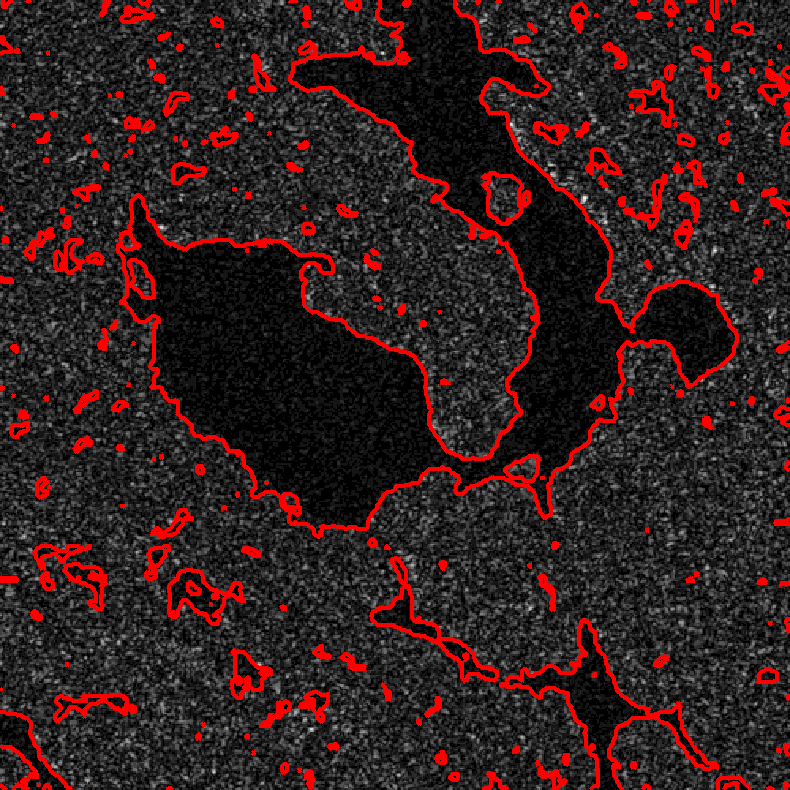}
\caption{Illustration of segmentation results by classical active contour~\cite{Shuai} with different parameters, from left to right: $\lambda=0.4, 0.3, 0.2, 0.1$, where $\lambda$ controls the weight of regularization.}
\label{fig:activecontour}
\end{figure*}

As is often the case, SAR images suffer from speckle effects and non-local radiometric variations (changes of backscattering coefficients) in background and objects. The speckle effect probably introduces false alarm, while the non-local variations of backscattering coefficients introduce missing detections. One may argue that these problems can be handled by choosing a good balance parameter $\lambda$ in the active contour model.
However, note that the false alarms raised by speckle effects need a larger $\lambda$, but the missing detections resulted by non-local radiometric variations demand a smaller $\lambda$ on the contrary, which are hard to achieve. Fig.~\ref{fig:activecontour} shows the segmentation results of a classic active contour model~\cite{Shuai} with different parameters $\lambda$. We can see, if $\lambda$ is large, some details of the object is lost, however, if $\lambda$ is small, the results have a lot of false alarms. It seems to be impossible to obtain an accurate result from parameters regulating.

In order to solve this, the NLAC method segments the regions by computing the similarity between each patch and its non-local neighbor patches. As modeling patches is more robust to model pixels in terms of speckle, NLAC seems to be more robust than the classical method. Besides, NLAC separates the probability model of SAR image from the algorithm, which means that we only need to calculate the distribution metrics of patches by modeling the patches with probability mass function at once. Due to this, it is very convenient to choose a suitable candidate probability mass function to model each kind of SAR images.
More importantly, NLAC propagates non-locally, which means local variance have less impact on the final segmentation results, which is the reason why this NLAC segmentation algorithm can handle the global radiometric variations in the images.

\subsection{Does the multi-scale strategy help the NLAC?}
In this section, we present how the multi-scale strategy can help the NLAC method. We experiment on a simulated SAR image, which is generated by adding multiplicative noise to a pure image. For the sake of comparison with the algorithm in \cite{Shuai}, where gamma distribution is applied to model SAR images for the segmentation, we also simulate a SAR image via gamma distribution given in Tab.~\ref{tab:distribution}.
The simulated SAR image is illustrated in Fig.~\ref{fig:syn_seg}, by the left, with the shape parameter $\alpha=4$ and the scale parameter $\beta=1$ of the gamma distribution. The size of the synthetic SAR image is $512\times 512$. It consists of three patterns: a double-circle, a triangle, and a horseshoe-shaped object with its intensity changing gradually.

Under the distribution prior, we choose gamma distribution to model image patches and use the Kullback-Leibler (KL) divergence to measure the similarity between patches. As illustrated in Fig.~\ref{fig:flowchart}, we apply the multi-scale NLAC method to the simulated SAR image. In this experiment, the size of patch is set to be $5$, the non-local size is set to $30$ and the weight $\lambda$ is set as $20$.

\subsubsection{Speedup of the convergence}
\begin{figure}[htb!]
\centering
  \includegraphics[width = 0.7 \linewidth]{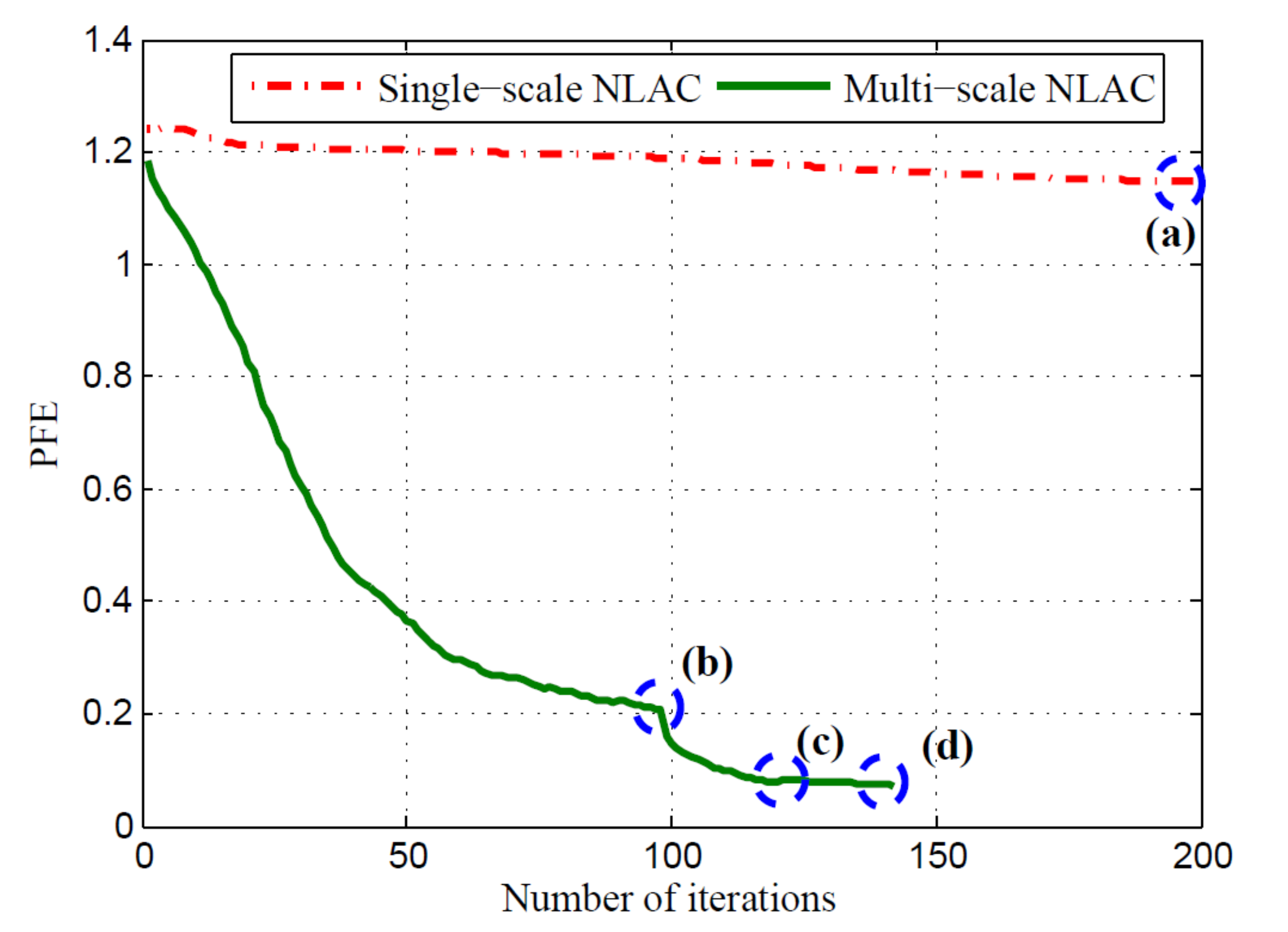}
  \caption{The comparison of the convergence speed between single-scale NLAC algorithm and the multi-scale NLAC (MS-NLAC) algorithm. The horizontal axis denotes the iterative times, while the vertical axis denotes the RFE defined in Eq.~\eqref{eq:accurate}. The red line is the convergence curve of single-scale NLAC, while the green one is the convergence curve of multi-scale NLAC.}
  \label{fig:syn_convergence}
\end{figure}

Figure~\ref{fig:syn_convergence} compares the convergence performance of the proposed MS-NLAC algorithm and the single-scale NLAC algorithm. Note that, for a fair comparison, we set the parameter of the single-scale NLAC to be the same with the multi-scale NLAC.
According to Figure~\ref{fig:syn_convergence}, we can see that MS-NLAC converges with less iterations than that of NLAC. In order to interpret this phenomena, it is worth noting that the energy minimization of the NLAC algorithm is a non-convex problem, and there exist many local minimum in the solution space.
Though Eauqtion~\eqref{eq:varphi_i} derives an efficient way to approach a feasible solution, it is still time-consuming and depends a lot on the initialization.
While, under the multi-scale strategy, at each step the NLAC model starts at a better initialization (i.e. segmentation from a coarser scale), which thus speed up the convergence of the algorithm.

\subsubsection{Improvement of segmentation precision}
\begin{figure*}[htb!]
\centering
  \includegraphics[width=0.24 \linewidth]{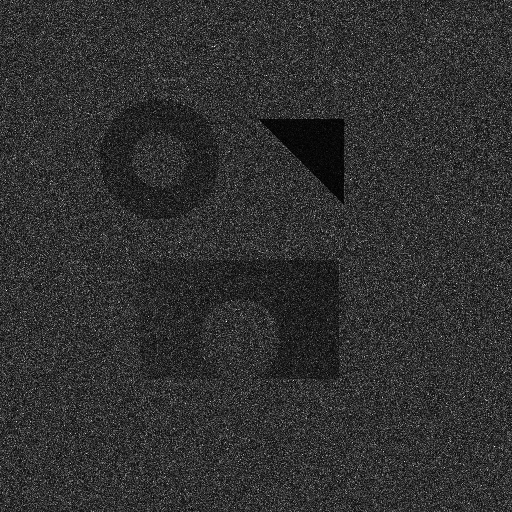}
  \includegraphics[width=0.24 \linewidth]{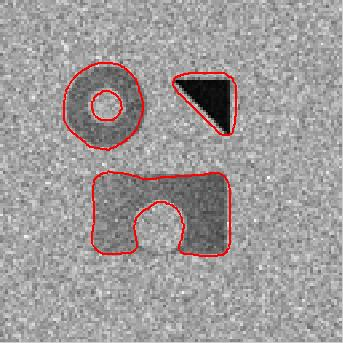}
  \includegraphics[width=0.24 \linewidth]{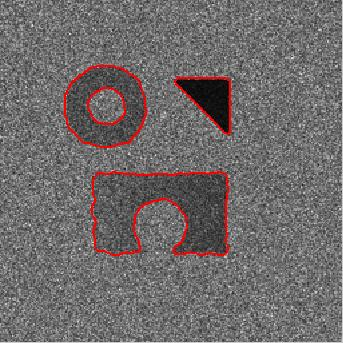}
  \includegraphics[width=0.24 \linewidth]{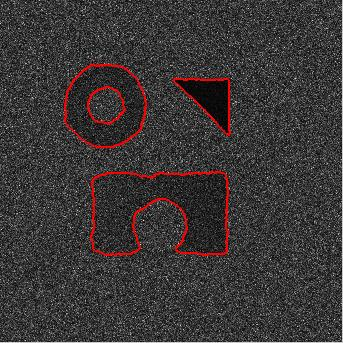}
  \caption{The segmentation results at each scale of the proposed MS-NLAC method. \emph{\textbf{From left to right}}: the synthetic SAR image; the segmentation result at scale $f^{[2]}$ with size of $128\times 128$; the result at scale $f^{[1]}$ with size of $256\times 256$; the finally segmented result at the original scale $f^{[0]}$ with size of $512\times 512$.}
  \label{fig:syn_seg}
\end{figure*}

From Figure.~\ref{fig:syn_convergence}, besides the convergence speed, one can also observe that MS-NLAC finally approaches much lower RFE than NLAC does.
Instead of segmenting the image at a fixed scale, MS-NLAC first finds a solution at a coarse scale and then refines it step-by-step in the scale space. This refinement process can be checked via the decreasing of RFE measure in Fig.~\ref{fig:syn_convergence}, where the blue circles indicate the results at different scales.
The segmentation results corresponding to these blue circles are displayed in Figure~\ref{fig:syn_seg}.
Note that the proposed MS-NLAC method can segment out all the objects at different scales, and more accurate results are achieved at finer scale. While, the single-scale NLAC hardly converges to the border of objects in this experiment.

\begin{figure*}[htb!]
\centering
  \includegraphics[width=0.24 \linewidth]{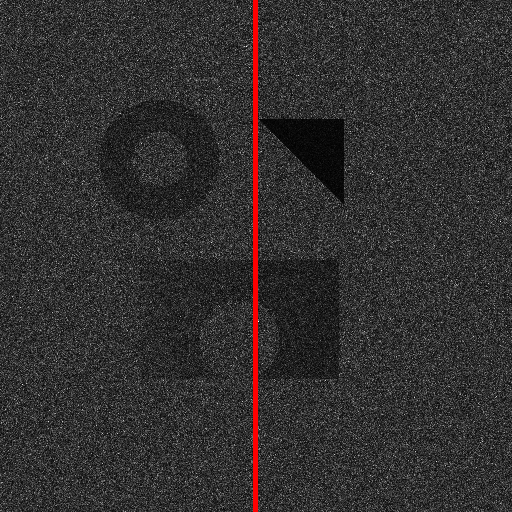}
  \includegraphics[width=0.24\linewidth]{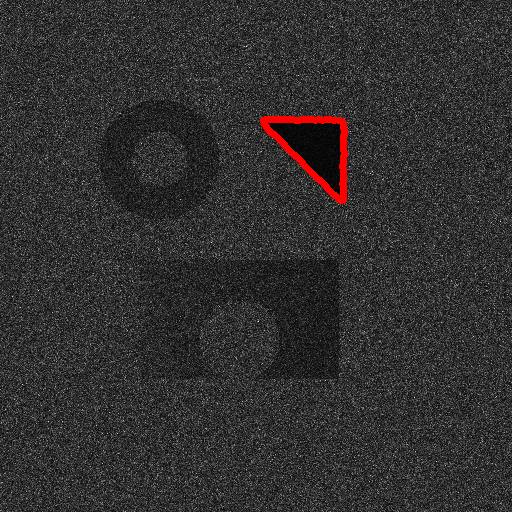}
  \includegraphics[width=0.24 \linewidth]{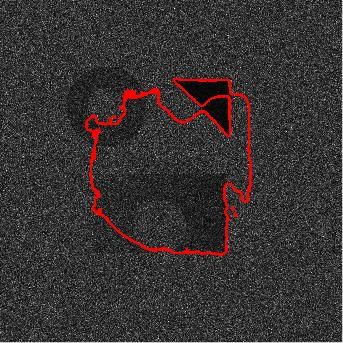}
  \includegraphics[width=0.24 \linewidth]{./syn1_seg.png}
  \caption{The comparison results. \textbf{\emph{From left to right}}: the result achieved by the method in~\cite{Shuai}; the result achieved by the method in~\cite{Marques}; the result using single-scale NLAC and the result of the proposed multi-scale NLAC method.}
  \label{fig:syn_seg_other}
\end{figure*}

The comparisons with the method in~\cite{Shuai} and \cite{Marques} are also displayed in Fig.~\ref{fig:syn_seg_other}.
The method in~\cite{Shuai} can hardly process this image and it even does not form a meaningful segmentation result. However, the method in \cite{Marques} can only extract out one object in the three\footnote{The segmentation result on this image by the method in~\cite{Marques} is kindly provided by its authors of the paper.}.

According to the experimental results, we can now conclude that the proposed multi-scale NLAC method outperforms the single-scale one: it not only propagates the contour much faster and can also converge to some states with lower energy, which leads to more accurate segmentation results.
It is also clearly from this experiment that the multi-scale NLAC method works better than those ones in~\cite{Shuai} and in ~\cite{Marques}. In next section, we will validate our method by utilizing real SAR images.

\subsection{Results on real SAR images}
\begin{figure*}[htb!]
\centering
  \includegraphics[width = 0.32\linewidth]{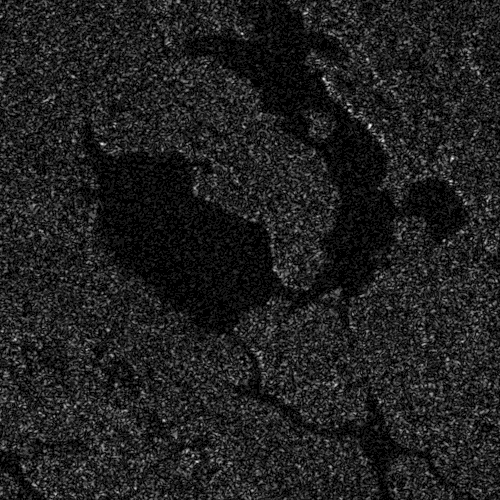}
  \includegraphics[width = 0.32\linewidth]{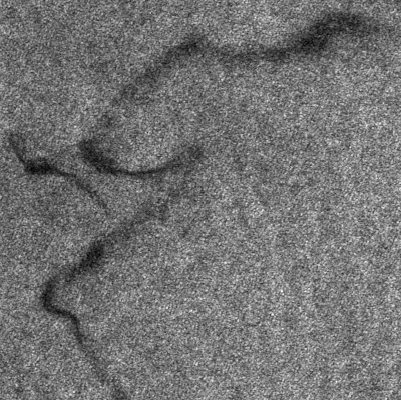}
  \includegraphics[width = 0.32\linewidth]{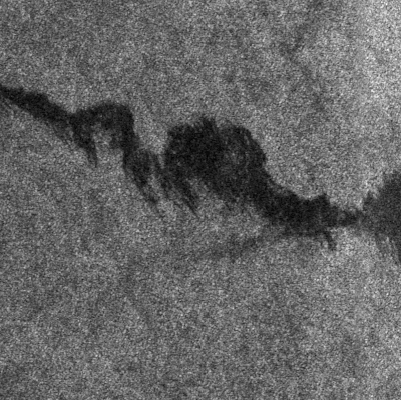}
  \caption{Real SAR images used for testing our algorithm. \emph{\textbf{From left to right}}: a scene of pond from an X-band Cosmo-Skymed SAR image and two sub-images containing oil spill from a C-band ENVISAT ASAR image.}
  \label{fig:sar}
\end{figure*}

In this section, we test the capability of the proposed algorithm by applying it on different kind of SAR images, shown in Fig.~\ref{fig:sar}:  a $500\times 500$ sub-image about a scene of pond from an X-band Cosmo-Skymed SAR image, with a spatial resolution of $3$ meters; and two $400\times 400$ sub-images (sub-image $1$ and sub-image $2$) about the oil spill from a C-band ENVISAT ASAR image with a spatial resolution of $30$ meters.

As mentioned before, in our algorithm, patch can be modeled by different probability density function in Tab.\ref{tab:distribution} and patch similarity could be measured by several distance in Tab.\ref{tab:distance}. In our experiments, we evaluate the combination of different distributions and distance respectively for each SAR image with the quantitative segmentation results. We also compare our results quantitatively with those achieved by the methods in~\cite{Shuai} and~\cite{Marques}. For the method~\cite{Shuai}, the parameters are tuned to obtain its best results.

For a fair comparison, in our algorithm, we set the same parameters to all the experiments. Here, the half patch size is set to be $w=7$, the non-local size is set to be $q=61$, the weight is set to $\lambda=20$, the number of scales is set to be $L=3$ and the iteration stop parameter $\omega=10^{-3}$.

\paragraph{Experiment on an X-band Cosmo-Skymed SAR image}
\label{sec:exp_X-band}
Tab.\ref{tab:1} provides the RFEs of the MS-NLAC algorithm for all the possible combination of distributions and metrics given in Tab.\ref{tab:distribution} and Tab.\ref{tab:distance}, respectively.
According to Tab.\ref{tab:1}, one can observe that:
\begin{itemize}
\item[-] The proposed MS-NLAC algorithm depends more on the patch similarities, rather than on the probabilistic models of patches. With the same patch models, KL-divergence $D_{KL}$ outperforms the others. One possible explanation is that KL-divergence is calculated by the ratio of probability density functions directly, which can reduce the impact of speckle in SAR images.
\item[-] With the same patch similarities, the proposed MS-NLAC algorithm archives the best performance when using log-normal distribution for describing patches. While, Gamma distribution provides comparable results. This in fact relates to the complexities of patch models. Though comprehensive probabilistic models, e.g. Weibull distribution and $G_A^0$ distribution, can model patches with more precision, it involves more parameters and the estimations of them are not stable by relying only on a small set of pixels, say, a patch with size of $7 \times 7$.
\end{itemize}

\begin{figure*}[htb!]
\centering
  \includegraphics[width=0.32 \linewidth]{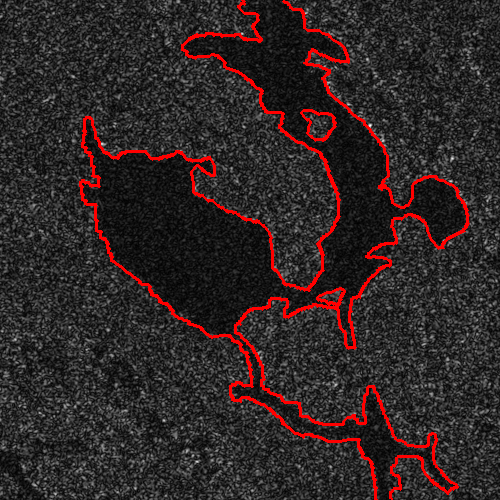}
  \includegraphics[width=0.32 \linewidth]{./segs1.png}
  \includegraphics[width=0.32 \linewidth]{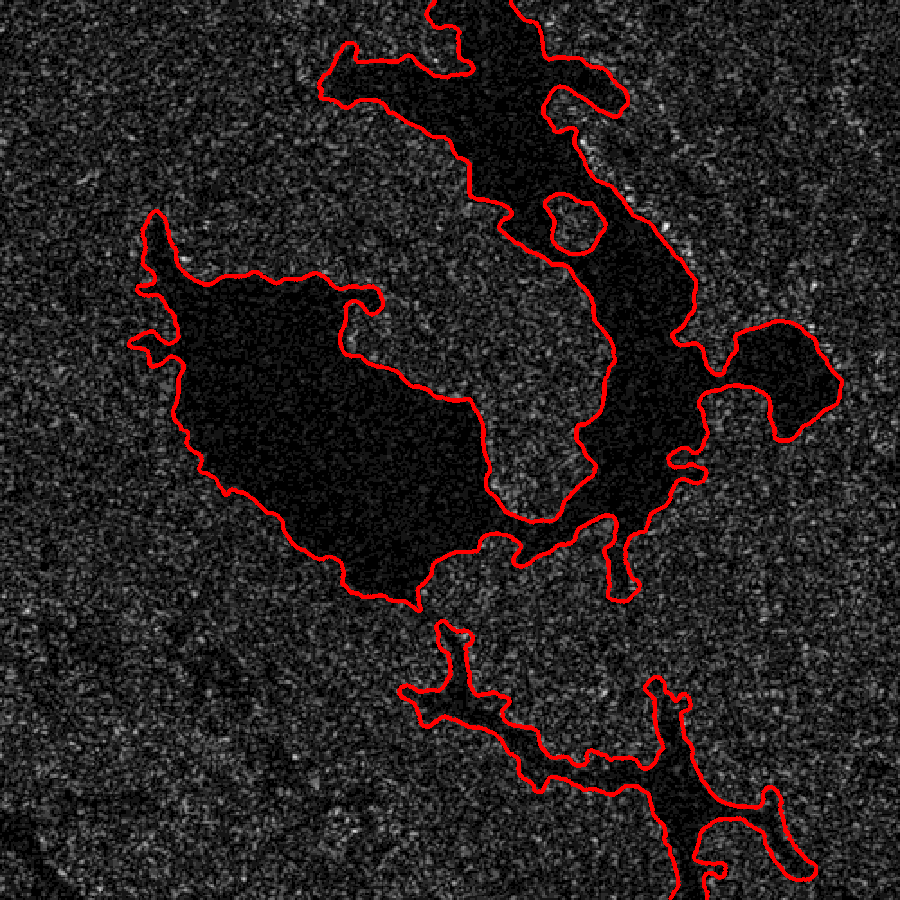}
  \caption{The results on an X-band Cosmo-Skymed SAR image. \emph{\textbf{From left to right}}: the ground truth, the results obtained by the method in~\cite{Shuai} and the result of the MS-NLAC algorithm with KL-divergence and log-normal distribution. Refer to the texts for more details.}
  \label{fig:KL9}
\end{figure*}

Fig.~\ref{fig:KL9} illustrates the segmentation result achieved by the MS-NLAC algorithm, where the boundaries of segmented objects are overlayed on the original image. The segmentation groundtruth of this image, in the left of Fig.~\ref{fig:KL9}, is provided by the authors of~\cite{Shuai}. Our result displayed on the right of Fig.~\ref{fig:KL9} is obtained by using log-normal distribution and KL-divergence, getting an RFE score $0.1231$. While, the method in \cite{Shuai}, utilizing conventional active contour model and Gamma distribution, achieves a segmentation result with an RFE score as $0.1863$, which is worse than that obtained by our MS-NLAC algorithm with its RFE score as $0.1393$.
By checking the segmentation regions in details, one can see that some undesired regions are detected and some parts of the pond are missed in the result achieved by \cite{Shuai}. While, our MS-NLAC algorithm can segment all the pond regions and the boundaries are more accurate, \emph{i.e.} the boundaries are more consistent with the outline profiles of the objects.

%


\begin{table}[htb!]
  \centering
  \caption{The results of RFE from different distributions and different metrics tested on the $500\times 500$ sub-image about a scene of pond from the X-band Cosmo-Skymed SAR image}
  \begin{tabular}{c|ccccc}
  \hline
  {}    & Log-normal & Rayleigh & Gamma & Weibull & $G_A^0$ \\ \hline
  $D_{KL}$    & \textbf{0.1231}   &  0.2139 & \textbf{0.1393} & 0.2109 &0.2701\\\hline
  $D_{JS}$    & 0.8058  & 0.8549  & 0.9541 & 0.9378 &1.3193\\\hline
  $D_{TV}$    & 1.7693  & 1.9042  & 1.7907 & 1.7948 &1.8881\\\hline
  $D_{EM}$    & 1.6774 & 1.8850  & 1.6828 & 1.6825 &1.8111\\\hline
  $D_{H}$ & 1.7693 &1.9111 &1.7959  &1.8052 &1.8796\\\hline
  \end{tabular}
  \label{tab:1}
\end{table}

\paragraph{Experiment on C-band ENVISAT ASAR images}
The objective of this experiment is to use the MS-NLAC algorithm to extract oil-spill regions over the sea from SAR images. The main difficult of this task lies in the fact that the waves of the sea often introduce some false detections and the boundaries of the oil-spill regions are ill-defined. What's more, the non-local radiometric variations also have a subtle influence impact on the segmentation task, as explained in Fig.\ref{fig:SAR_profiles}.

\begin{figure*}[htb!]
\centering
  \includegraphics[width=0.32 \linewidth]{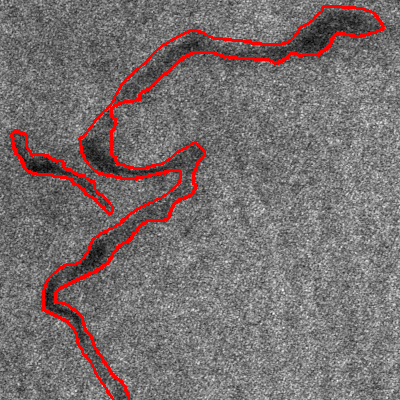}
  \includegraphics[width=0.32 \linewidth]{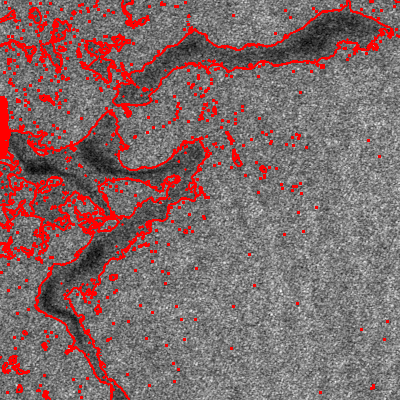}
  \includegraphics[width=0.32 \linewidth]{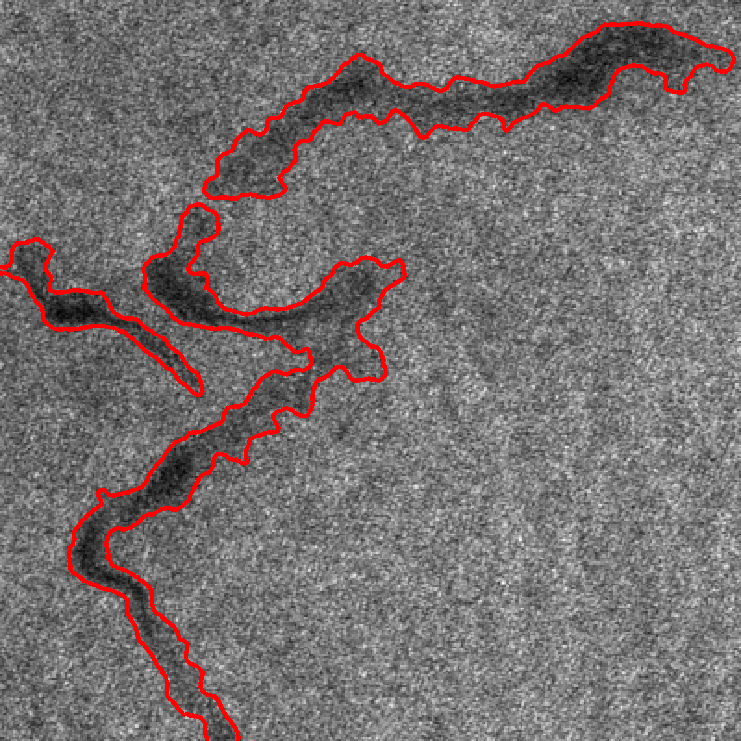}  \\
\vspace{1mm}
  \includegraphics[width=0.32 \linewidth]{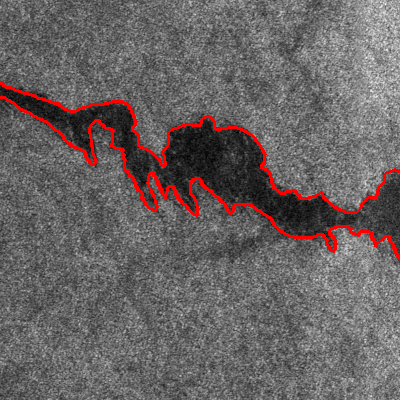}
  \includegraphics[width=0.32 \linewidth]{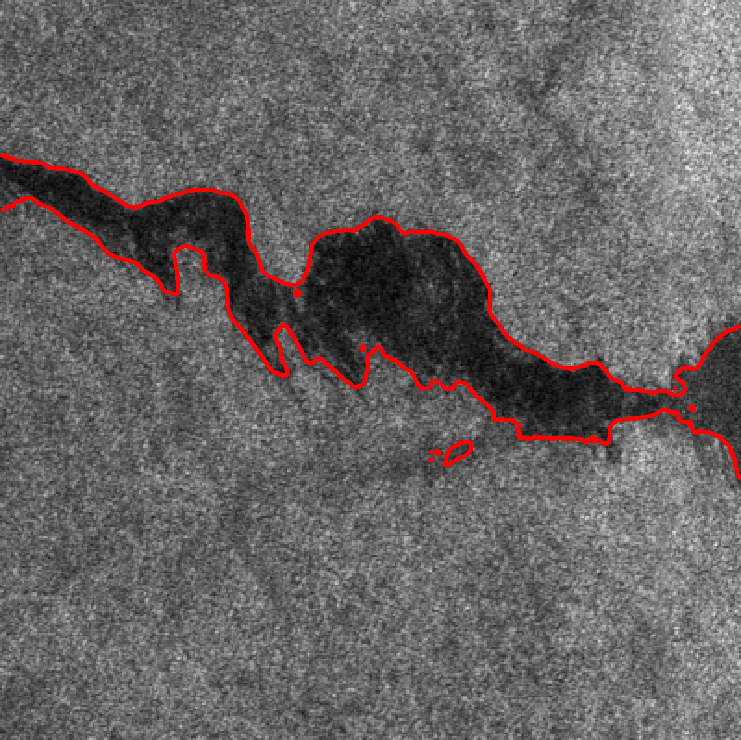}
  \includegraphics[width=0.32 \linewidth]{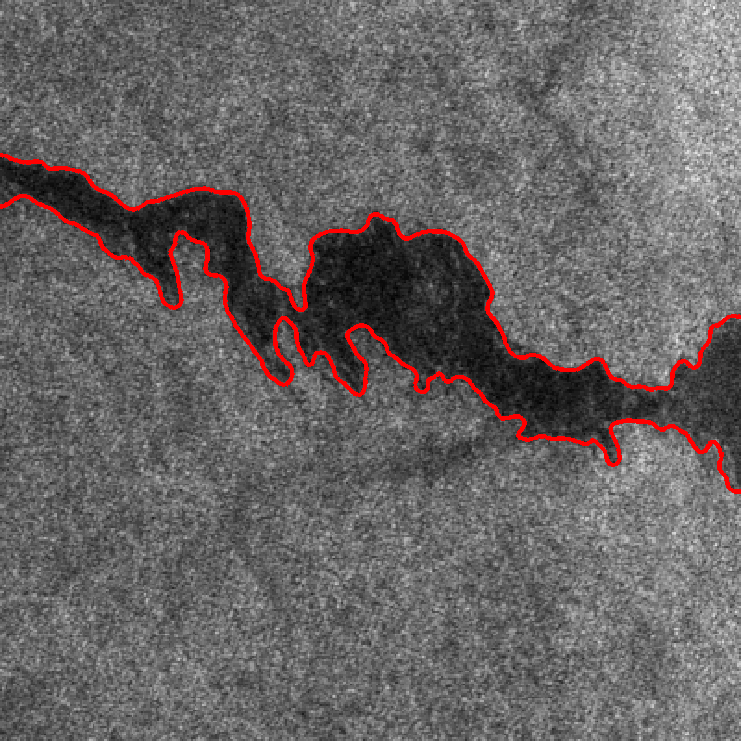} \\
  \caption{The segmented results of two $400\times 400$ sub-images about the oil spill from C-band ENVISAT ASAR image. \emph{\textbf{From left to right}}: the ground truth, the results obtained by the method in~\cite{Shuai} and the result of the MS-NLAC algorithm with KL-divergence and log-normal distribution. Refer to the texts for more details.}
  \label{fig:kl2}
\end{figure*}

\begin{table}[htb!]
  \centering
  \caption{The RFEs of segmentation results achieved by MS-NLAC algorithm with different distributions and metrics tested on the $400\times 400$ subimage-$1$ about a scene of oil spills in a C-band ENVISAT ASAR image.}
  \begin{tabular}{c|ccccc}
  \hline
  {}    & Log-normal & Rayleigh & Gamma & Weibull &$G_A^0$\\ \hline
  $D_{KL}$   & \textbf{0.3648}   &  1.2458 & \textbf{0.4348} & 0.9643 &0.5749\\\hline
  $D_{JS}$   & 0.4370  & 0.4372  & 0.4882 & 0.4967 &0.4635\\\hline
  $D_{TV}$   & 4.4473  & 5.4266  & 5.1523 & 5.8919 &3.7927\\\hline
  $D_{EM}$   & 5.0348  & 5.6654  & 5.4835 & 6.4862 &3.5155\\\hline
  $D_{H}$    & 5.4807 &5.1492 &5.1106  &5.4903 &3.6857\\\hline
  \end{tabular}
  \label{tab:2}
\end{table}

\begin{table}[htb!]
  \centering
  \caption{The RFEs of segmentation results achieved by MS-NLAC algorithm with different distributions and metrics tested on the $400\times 400$ subimage-$2$ about a scene of oil spills in a C-band ENVISAT ASAR image.}
  \begin{tabular}{c|ccccc}
  \hline
  {}    & Log-normal & Rayleigh & Gamma & Weibull &$G_A^0$\\ \hline
  $D_{KL}$   &  \textbf{0.1143}  & 1.2803 & \textbf{0.1175} & 3.4178 &2.1103\\\hline
  $D_{JS}$   & 0.4264  & 0.2329  & 0.2843 & 0.2655 &0.7811\\\hline
  $D_{TV}$   & 3.8689  & 4.9299  & 4.2406 & 4.3007 &4.5505\\\hline
  $D_{EM}$   & 3.5814  & 5.0301  & 3.1785 & 3.7486 &3.7327\\\hline
  $D_{H}$ & 3.8450 &4.2687 &4.3963  &4.6569 &3.9730\\\hline
  \end{tabular}
  \label{tab:3}
\end{table}

By using the MS-NLAC algorithm, Tab.\ref{tab:2} and Tab.\ref{tab:3} give the quantitative segmented results of all the combination of different distributions and metrics for these two sub-images respectively, from which we can have similar observations as those on the experiment in previous section:
\begin{itemize}
\item[-] In the MS-NLAC algorithm, the patch similarities play a more important role than that of the probabilistic patch models.
    Comparing each column of the Tab.\ref{tab:2} and Tab.\ref{tab:3}, one can see that KL-divergence $D_{KL}$ and Jensen-Shannon (JS) divergence~\cite{Lin1991} $D_{JS}$ achieve comparable results but outperform the others. Besides the advantages on calculations~\cite{Deledalle2015}, these two divergences are more suitable on measuring the similarity between two probability distributions.
\item[-] The best performances are also achieved by using log-normal distribution as probabilistic patch models, by utilizing the same patch similarities. This is consistent with the observations in~\cite{Delignon97}, where empirical log-normal distribution showed superiority on describing ocean SAR images.
\end{itemize}

The comparisons between our MS-NLAC algorithm and the algorithm in~\cite{Shuai} are displayed in Fig.~\ref{fig:kl2}, where the segmentation groundtruth are annotated by SAR image interpretation expert.
When using Gamma distribution as probabilistic patch models, the MS-NLAC algorithm achieves the segmentation RFEs as $0.5248$ and $0.1429$ on the subimage-1 and subimage-2 respectively, outperforming the results obtained by the approach in~\cite{Shuai} with the RFEs as $0.4348$ and $0.1175$.

The results achieved by the method~\cite{Shuai} are displayed in the middle of Fig.~\ref{fig:kl2}. The result on subimage-1 is acceptable except small false detected regions and some inaccurate boundaries, while the false detection rate on subimage-2 is very high, as many small regions are obtained by mistake.
The right of Fig.~\ref{fig:kl2} shows the segmented results obtained by MS-NLAC algorithm with KL divergence and log-normal distribution.
It obtains better results for these two SAR images, especially the boundaries of the oil spills are with high precision.

Another state-of-the-art approach is the method in~\cite{Marques}, but it can not produce meaningful segmentation results on these two images\footnote{We provided our original SAR images to the authors of~\cite{Marques}. N. J. Marques, one of the author, run their algorithm on those images and told us the algorithm could not segment those two SAR images.}.
Thus, concluding these experiments, the proposed method shows its capability of handling the problems of segmenting meaningful objects from SAR images.

\section{Conclusion}

This paper addressed the problem of meaningful object segmentation from SAR images, which often suffers from the speckles and non-local variations of backscattering coefficients. In contrast with existing works on this topic, that concentrate on eliminating the speckle effects to the image segmentation, we suggested to simultaneously consider the two challenges when extracting meaningful objects from SAR images, by investigating the active contour model with non-local principles.
More precisely, we first introduced and adapted the non-local active contour model, \emph{i.e.} NLAC, to the problem of SAR image segmentation, by measuring local patch similarities and integrating the patch interactions in a non-local way. In order to lead the NLAC algorithm fast converging to a better solution, we proposed a multi-scale non-local active contour method, so-called MS-NLAC algorithm. By experimenting on both simulated and real SAR images from different sensors, we discussed the necessity and efficiency of the MS-NLAC algorithm. It has demonstrated the superiority of our MS-NLAC algorithm on segmenting meaningful objects, e.g. oil spills, from SAR images with heavy speckles and non-local variations of backscattering coefficients.

It is also interesting to observe that the MS-NLAC algorithm depends more on the patch similarities rather than on probabilistic patch models. Thus, this work also provides a flexible way to incorporate different descriptive models of patches and similarity measures between them in the segmentation problem.
In our current work, we demonstrated the use of several patch models and similarities, but further investigating on them are also of great interest. For instance, one could combine the MS-NLAC algorithm with the comprehensive studies of non-local patch similarities in~\cite{Deledalle2014, Deledalle2015} and extend it to the partition of polarimetric SAR images.

\section*{Acknowledgment}
The authors would like to thank Y. Shuai for sharing his codes in~\cite{Shuai} and thank N. J. Marques for providing his segmentation results on our images by using his method proposed in~\cite{Marques}.
This research is supported in part by the National Natural Science Foundation of China (No.91338113 and No.61271401) and the National Key Basic Research and Development Program of China under contract 2011CB707105.

\bibliographystyle{IEEEtran}
\bibliography{reference}
\end{document}